\documentclass[11pt]{article}
\usepackage[ letterpaper, top=0.85in, bottom=0.85in, left=0.90in, right=0.90in]{geometry}
\usepackage[T1]{fontenc}
\usepackage{newtxtext}
\usepackage{booktabs}
\usepackage{multirow}
\usepackage{array}
\usepackage{tabularx}
\usepackage{threeparttable}
\usepackage{siunitx}
\usepackage{enumitem}
\usepackage{microtype}
\usepackage{xcolor}
\usepackage{mathtools}
\usepackage{amssymb}
\usepackage{newtxmath}
\usepackage{titlesec}
\titlespacing*{\section}
{0pt}
{1.9ex plus 0.4ex minus 0.2ex}
{0.75ex plus 0.2ex}
\titlespacing*{\subsection}
{0pt}
{1.45ex plus 0.3ex minus 0.2ex}
{0.50ex plus 0.2ex}
\titlespacing*{\subsubsection}
{0pt}
{1.20ex plus 0.3ex minus 0.2ex}
{0.40ex plus 0.2ex}
\setlist[itemize]{topsep=4pt, itemsep=2pt, parsep=0pt, partopsep=0pt}
\setlist[enumerate]{topsep=4pt, itemsep=2pt, parsep=0pt, partopsep=0pt}
\usepackage{tikz}
\usetikzlibrary{arrows.meta, positioning, fit, calc, shapes.geometric, backgrounds}
\usepackage{pgfplots}
\pgfplotsset{compat=1.18}
\usepackage{graphicx}
\usepackage{subcaption}
\usepackage{float}
\usepackage{placeins}
\usepackage{xurl}
\usepackage[numbers, sort&compress]{natbib}
\usepackage[hidelinks, breaklinks=true]{hyperref}
\providecommand{\Description}[1]{}

\begin{document}
\title{
    \textbf{VERA-8B: Evidence-Grounded Audit Risk Reasoning from SEC Filings}
    \\[0.35em]
    \large A Unified Audit-Evidence Standard for Prospective Detection and Selective Review
}
\author{
    Menghan Liu
    \\
    New York University
    \\
    New York, NY, USA
    \\
    \texttt{ml9010@nyu.edu}
    \and
    Elynn Chen\thanks{Corresponding author.}
    \\
    New York University
    \\
    New York, NY, USA
    \\
    \texttt{elynn.chen@stern.nyu.edu}
}

\date{}
\maketitle
\begin{abstract}
Across audit applications, judgments must be supported by reasonable evidence. However, standard financial language models prioritize fluency over evidence. They are built for general financial reasoning and may produce plausible but ambiguous answers, creating a grounding gap that makes them unsuitable for audit work. We address this gap with VERA-8B, a new end-to-end audit reasoning system that identifies audit risks before enforcement actions occur. Constructing such a model raises several challenges, as no prior machine learning work targets pre-enforcement audit prediction. To our knowledge, we are the first to unify SFT and GRPO for evidence-grounded audit reasoning under one evidence standard, achieving performance that surpasses all evaluated baselines. Because auditing cannot tolerate unsupported claims, we introduce abstention and uncertainty qualification to defer uncertain or evidence-incomplete cases. Finally, we design an AuditBridge to ground model reasoning for practical audit work. It transforms raw filings into verified records and then into reviewer-ready reports, bridging finance and computation with broad generality. Together, these components produce auditable, review-ready outputs suitable for practical audit work.
\end{abstract}
\vspace{0.30em}
\vspace{0.65em}
\section{Introduction}
\label{sec:introduction}
Audit reasoning requires evidence, not simply plausible prediction. SEC filings contain abundant audit-related language, but lexical presence alone does not make it evidence. For example, a material-weakness disclosure may describe a current deficiency, a remediated history, or mere boilerplate. The same ambiguity also arises across other audit-relevant disclosures, including internal-control weaknesses, going-concern disclosures, and accounting estimates \cite{pcaob2201,pcaob2415,pcaob2501}. Hence, a red-flag word is not evidence. An exact quote carries evidential weight only if its timing, context, and audit mechanism support the claim.

Even with machine learning, core audit-evidence problems remain unresolved. Misstatement detection models deliver disciplined firm-level predictions but lack the exact filing evidence that supports a specific audit mechanism \cite{dechow2011misstatements,bao2020fraud,bertomeu2021misstatements,zavitsanos2021realistic}. Although financial language models produce richer representations and explanations \cite{yang2020finbert,chen2021finqa,liu2025finr1,qian2025fino1},
they can still cite absent spans, assign valid quotations to the wrong category, or treat boilerplate as direct support. Recent audit benchmarks and expert-guided systems broaden audit reasoning and evidence discovery \cite{wang2025finauditing,liu2026auditfraudbench,bai2025auditagent}. For example, AuditAgent \cite{bai2025auditagent} uses enforcement documents and financial reports to locate fraud evidence, whereas VERA-8B predicts audit risk from the filing alone before enforcement occurs. More broadly speaking, none of them govern the risk decision, evidence provenance, rationale, and abstention by a single standard.

The training labels create a second problem. Enforcement actions occur after the filing, while audit prediction must rely only on information available before enforcement \cite{secedgar,secaaer,zavitsanos2021realistic}. A later enforcement action tells us which issuer--period deserves investigation, but not what evidence was available in the filing at the time. Likewise, the absence of an enforcement action does not mean that the filing contains no audit risk. Enforcement outcomes therefore cannot be used as direct filing-level evidence labels.

\begin{figure}[!t]
    \centering
    \includegraphics[width=\linewidth]{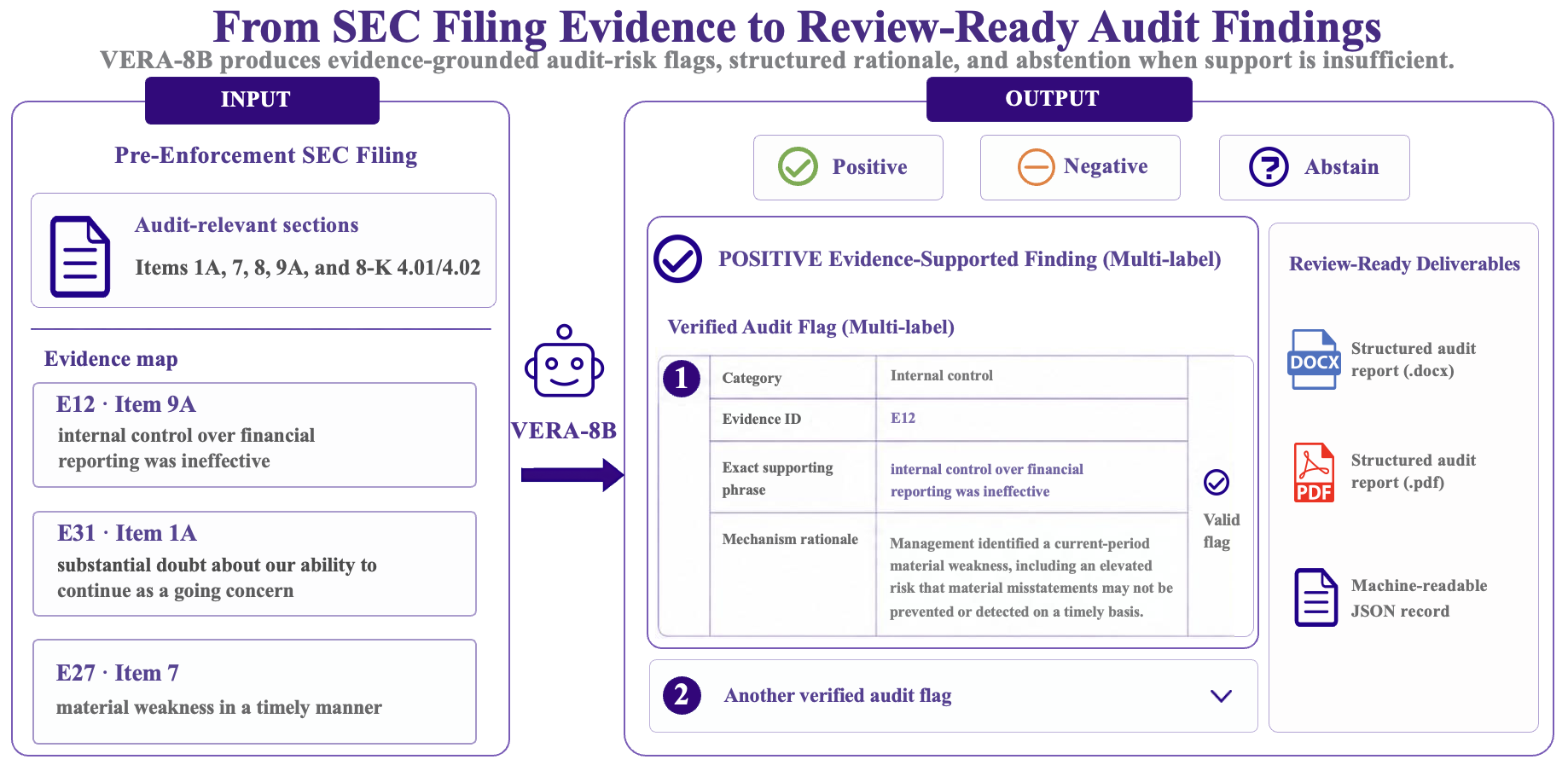}
    \caption{VERA-8B converts SEC filings into auditable outputs.}
    \Description{
    Overview of VERA-8B input and output. The left side shows
    pre-enforcement SEC filing evidence. The right side shows
    multi-label evidence-supported audit findings, structured report
    fields, and review-ready deliverables.
    }
    \label{fig:intro-overview}
\end{figure}

We address these gaps with VERA-8B, to our knowledge the first audit language model that reasons about future risks directly from filing evidence. VERA-8B goes beyond prediction by producing a complete audit finding. It identifies the risk, explains the reasoning with exact
filing evidence, and abstains when support is insufficient. Figure~\ref{fig:intro-overview} illustrates this process, showing how our model transfers raw filing texts to evidence-backed, review-ready audit findings.

Overall, VERA-8B leads the benchmark on the full audit task. We evaluate it against both general-purpose and finance-specific language models under a frozen protocol. Two results stand out. First, VERA-8B delivers high accuracy while strictly grounding claims in evidence. It achieves 95\% binary F1, 88\% category micro-F1, 87\% evidence-verified micro-F1, and 94\% verified-positive recall, with valid structured outputs in all test cases and only 3\% unsupported claims. Second, these gains are large relative to the strongest external baseline for each measure. Binary F1 jumps by 45\% and unsupported claims drop by 81\%, with similarly strong improvements across all other metrics. In short, VERA-8B does more than flag risk. It produces auditable, evidence-backed findings that an auditor can review directly.
Our key contributions are as follows:
\begin{itemize}

    \item \textbf{A New Audit-Specific Reasoner with Benchmark-Leading Performance.} We introduce VERA-8B, the first end-to-end model purpose-built for prospective audit reasoning. It outperforms all general-purpose and finance-specific baselines on risk detection, multi-label reasoning, evidence-verified reasoning, and structured output.

    \item \textbf{A Unified Audit-Specific Post-Training Framework.} We design a post-training framework governed by a single evidence contract that covers supervision, generation, verification, and refinement. Structured SFT establishes the contract and verifier-guided adaptive GRPO tackles the remaining hard cases. This design is reusable across other evidence-intensive financial tasks.

    \item \textbf{Audit-Specific Reliability Control.} To ensure audit reliability, VERA-8B abstains when evidence is insufficient rather than forcing a prediction. We introduce exact evidence verification, explicit abstention, and uncertainty-qualified routing to defer unreliable cases.
    
    \item \textbf{AuditBridge: Machine-Readable and Reviewer-Ready Outputs.} We bridge reasoning to practice with AuditBridge. It converts filings into verified JSON records and renders them as reviewer-ready reports, forming a reusable pipeline from text to computation to review. This pipeline can also be applied broadly to other evidence-constrained tasks.

\end{itemize}

\section{Related Work}
Prior work has improved audit prediction, financial reasoning, and evidence discovery. However, gaps still remain as technology continues to evolve, and our model is designed to bridge these gaps.

\textbf{Misstatement Risk Detection Models.}
Accounting research has long estimated misstatement risk from financial ratios and firm-level variables, e.g., the Dechow F-score \cite{dechow2011misstatements}. Machine-learning approaches capture nonlinear patterns in raw accounting data \cite{bao2020fraud} and broader accounting, governance, market, and audit features \cite{bertomeu2021misstatements}, while text-based methods extract predictive signals from 10-K language \cite{loughran2011liability}. Realistic evaluation must account for class rarity, chronological splits, and delayed discovery to avoid inflated performance \cite{zavitsanos2021realistic}. However, even with this temporal design, predictions remain firm-level, lacking links to specific filing evidence and reasoning. These models can predict but not explain.

\textbf{LLM Post-Training and Selective Prediction.}
Firm-level prediction limitations motivate post-training methods for structured, controllable reasoning. LoRA and QLoRA provide parameter-efficient adaptation via low-rank updates and 4-bit quantization \cite{hu2022lora,dettmers2023qlora}. SFT learns desired response patterns from demonstrations, while GRPO refines
them through group-relative rewards
\cite{shao2024deepseekmath}. Temperature scaling, selective classification, and conformal prediction support uncertainty qualification by calibrating confidence and routing ambiguous cases to review \cite{guo2017calibration,geifman2017selective,angelopoulos2023conformal}. Selective abstention learning further teaches LLMs to defer when reliable support is unavailable \cite{huang2025abstention}. Together, these techniques cover adaptation, alignment, and uncertainty control, yet stay task-agnostic without audit-specific evidence rules.

\textbf{Financial LLMs and Task Specialization.}
Financial research applies these methods to domain-specific language and reasoning. FinBERT adapts representations to financial text, while FinQA evaluates numerical reasoning over financial reports \cite{yang2020finbert,chen2021finqa}. Fin-R1 and Fino1 further combine domain supervision with SFT and reinforcement learning \cite{liu2025finr1,qian2025fino1}. Their results confirm that targeted data and optimization improve financial reasoning across text, tables, and equations. Despite this progress, financial competence does not imply audit competence. Audit-specific benchmarks show that strong LLMs remain inconsistent across semantic, relational, and numerical auditing tasks \cite{wang2025finauditing}. However, these benchmarks do not directly assess whether a cited filing disclosure is current, unresolved, and sufficient to support a specific audit conclusion.

\textbf{Audit Reasoning and Multi-Label Evidence.}
Early audit-AI research developed rule-based expert systems for decision support in auditing \cite{hansen1982expert}, while subsequent work examined the value of explanations in audit expert systems \cite{ye1995explanation}. More recent work uses auditor-opinion text for prediction and expert-guided agents for cross-document fraud evidence discovery \cite{sideras2024bankruptcy,bai2025auditagent}. AuditFraudBench evaluates joint reasoning over filings, restatements, financial figures, disclosure narratives, and enforcement mechanisms \cite{liu2026auditfraudbench}, while multi-label learning supports several non-exclusive risks within one observation \cite{zhang2014multilabel}. However, existing work remains fragmented and is incapable of integrating these capabilities into a unified and standardized audit reasoning model.

Our research addresses these limitations through an evidence-constrained audit reasoning framework. It synergizes temporally aligned enforcement supervision, an executable rulebook, evidence-aligned multi-label SFT, audit-specific GRPO, and selective abstention. Together, they enable compact language models to make verifiable audit judgments and defer when evidence is insufficient.

\section{Task Definition and Data Sources}
\subsection{Audit-Risk Reasoning Task}
\label{sec:task-definition}

We formulate auditing as an evidence-grounded reasoning task over individual SEC filings. Given a pre-enforcement filing observation $x_i$, the model produces
\begin{equation}
f_{\theta}(x_i)
=
o_i
=
(\mathbf{y}_i,E_i,r_i,a_i),
\end{equation}
where $\mathbf{y}_i\in\{0,1\}^{9}$ is a multi-hot audit-risk category vector, $E_i$ contains supporting filing evidence, $r_i$ explains the evidence-to-risk relationship, and $a_i\in\{0,1\}$ denotes abstention.

The nine non-exclusive categories are internal control, restatement, revenue recognition, fraud or misconduct, going concern, covenant or default, disclosure opacity, auditor change or disagreement, and accounting estimates. An observation may contain zero, one, or multiple
categories.

Our task is to identify all filing-supported audit risk categories, cite the corresponding evidence, explain the underlying audit mechanism, and abstain when the available evidence is insufficient.

\subsection{Data Sources}

The framework draws on three source families with strictly separated roles: pre-enforcement filing information, ex-post enforcement outcomes, and public audit knowledge. Table~\ref{tab:data-sources} summarizes their notation and use.
\begin{table}[H]
\centering
\caption{Data Sources and Roles.}
\label{tab:data-sources}

\vspace{-3pt}
\includegraphics[width=\linewidth]{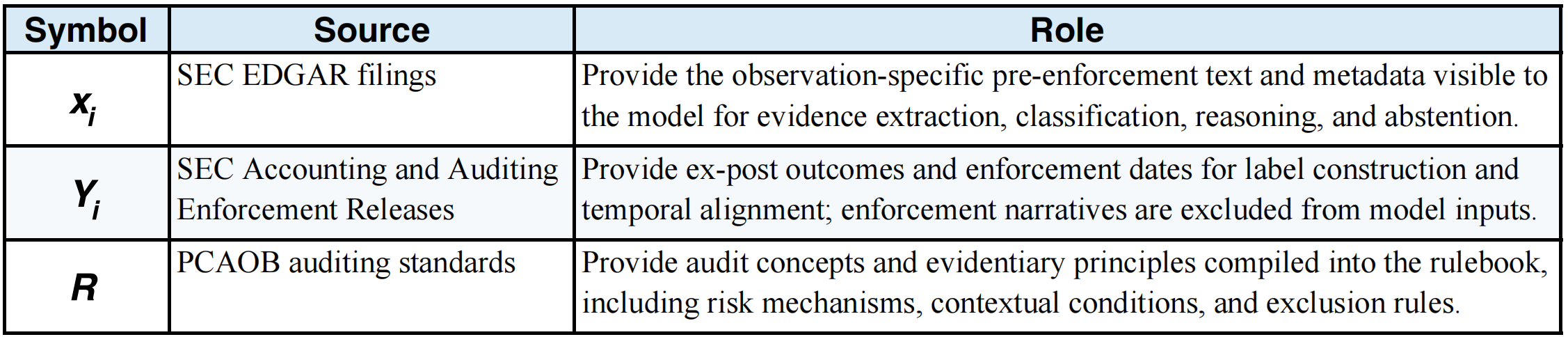}

\begin{minipage}{\columnwidth}
    \footnotesize
    \textit{Sources:}
    SEC EDGAR filings~\cite{secedgar};
    SEC Accounting and Auditing Enforcement
    Releases~\cite{secaaer};
    PCAOB auditing standards~\cite{pcaob2201,pcaob2401,pcaob2415,pcaob2501}.
\end{minipage}

\Description{
A three-column table summarizing the symbols, sources,
and strictly separated roles of SEC filings, enforcement outcomes,
and PCAOB auditing standards in supervision construction.
}

\end{table}

\begin{figure}[!t]
    \centering
    \includegraphics[width=\linewidth]{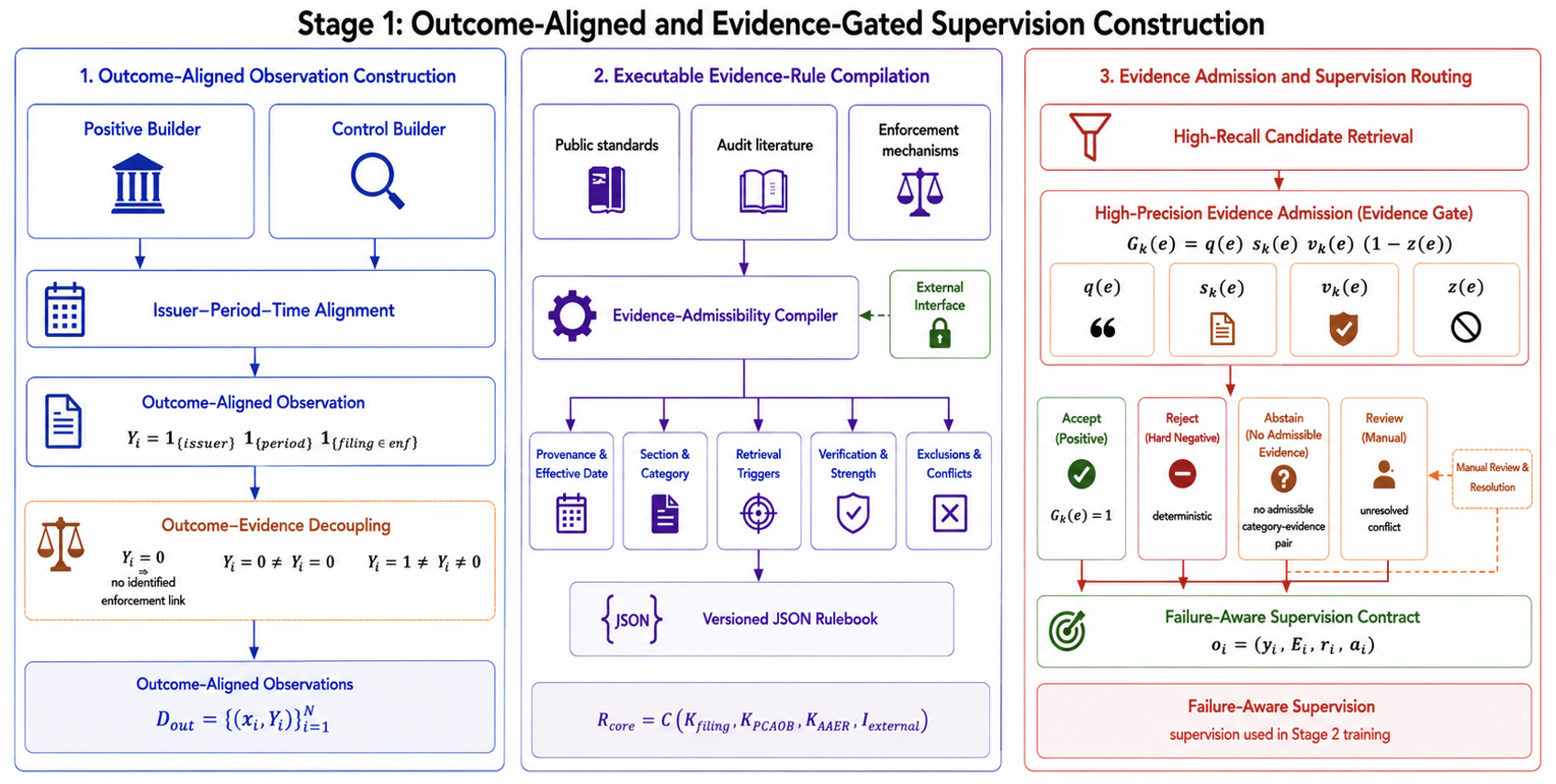}
    \caption{Stage~1 aligns filings with subsequent enforcement outcomes,
    compiles public audit knowledge into an executable rulebook, and
    converts filing text into evidence-gated supervision.}
    \Description{Stage~1 pipeline for outcome alignment, executable
    rulebook construction, and evidence-gated supervision.}
    \label{fig:stage1}
\end{figure}
\section{Methodology}
Our framework has two stages. Stage~1 determines what the model is allowed to learn from filing evidence. Stage~2 teaches and enforces that behavior through evidence-constrained post-training.
\subsection{Stage 1: Supervision Construction}
\label{sec:stage1}
Given the structured target
$o_i=(\mathbf{y}_i,E_i,r_i,a_i)$, Stage 1 constructs the supervision required to learn this mapping, as outlined in Figure~\ref{fig:stage1}. It proceeds in two steps:
\begin{equation}
x_i
\longrightarrow
(x_i,Y_i)
\longrightarrow
(\mathbf{y}_i,E_i,r_i,a_i),
\label{eq:supervision-pipeline}
\end{equation}
where $x_i$ is a firm--filing--fiscal-period observation and $Y_i$ is an ex-post enforcement-link indicator.
\subsubsection{Outcome-Aligned Dataset Construction}
\label{sec:data-construction}
The first step links filings to later enforcement outcomes without treating those outcomes as filing evidence. Because AAERs are event-level records that may span multiple entities, periods, and filings \cite{secaaer,dechow2011misstatements}, a filing is linked to an enforcement outcome only when issuer identity, affected fiscal period, and temporal precedence are jointly satisfied:
\begin{equation}
Y_i
=
\mathbf{1}_{\{M_i^{\mathrm{issuer}}=1\}}\,
\mathbf{1}_{\{M_i^{\mathrm{period}}=1\}}\,
\mathbf{1}_{\{d_i^{\mathrm{filing}}
<d_i^{\mathrm{enforcement}}\}}.
\label{eq:positive-definition}
\end{equation}
Uncertain issuer or period matches are retained for review or sensitivity analysis rather than treated as confirmed enforcement links.

Controls are sampled from the same SEC filing universe after excluding linked issuer--periods, known enforcement-linked issuers, and filings within the predefined contamination window. Positive and control observations use the same filing assembly and audit-relevant sections. We exclude post-enforcement narratives, split by issuer, and remove exact and near-duplicate passages across partitions.

Notably, enforcement status and admissible filing evidence remain separate. An enforcement link does not guarantee admissible pre-enforcement evidence, and its absence does not imply that the filing contains no audit-risk evidence. This asymmetry creates both evidence-supported positives and hard abstention cases. The resulting outcome-aligned dataset is
\begin{equation}
\mathcal{D}_{\mathrm{out}}
=
\left\{
(x_i,Y_i)
\right\}_{i=1}^{N}.
\label{eq:outcome-dataset}
\end{equation}
The next step determines which filing evidence can be admitted into category, reasoning, and abstention targets.
\subsubsection{Executable Rulebook Construction}
\label{sec:rulebook-construction}
The second step builds the executable audit-evidence standard that underpins VERA-8B. The rulebook converts public audit knowledge into a versioned JSON contract that maps filing evidence to audit-risk categories. Each rule captures its source, effective date, applicable filing section, risk category, retrieval trigger, verification conditions, exclusions, and minimum evidence strength. The compiler combines
\begin{equation}
R_{\mathrm{core}}
=
C\!\left(
K_{\mathrm{filing}},
K_{\mathrm{PCAOB}},
K_{\mathrm{AAER}},
I_{\mathrm{external}}
\right),
\label{eq:rulebook-compiler}
\end{equation}
where $K_{\mathrm{filing}}$ represents audit-relevant SEC filing sections, $K_{\mathrm{PCAOB}}$ captures public audit concepts and evidentiary principles from PCAOB
standards \cite{pcaob2201,pcaob2401,pcaob2415,pcaob2501}, and $K_{\mathrm{AAER}}$ abstracts recurring mechanisms from historical enforcement actions \cite{secaaer}. The interface $I_{\mathrm{external}}$ admits expert or firm-specific rules without weakening the evidence
standard.

The evidence gate separates retrieval from evidence admission. For a candidate
filing span $e$ and audit category $k$,
\begin{equation}
G_k(e) = q(e)\, s_k(e)\, v_k(e)\, \bigl(1-z(e)\bigr),
\label{eq:evidence-gate}
\end{equation}
where $q(e)$ requires an exact quotation, $s_k(e)$ checks section and context consistency, $v_k(e)$ verifies mechanism-level support, and $z(e)$ excludes historical-only statements and completed remediation. A candidate is admitted only when $G_k(e)=1$. Failed candidates become hard negatives, while unresolved conflicts are routed to manual review.

The compiler converts admitted evidence into training-ready targets. For each observation $i$,
\begin{equation}
o_i = \left( \mathbf{y}_i, E_i, r_i, a_i \right),
\label{eq:contract}
\end{equation}
where every positive category in $\mathbf{y}_i$ must pair with verified evidence $E_i$ and mechanism-level rationale $r_i$. If no admissible category--evidence pair exists, the observation receives an abstention target $a_i=1$. A later enforcement link alone can therefore never create a positive SFT target. As a result, Stage~1 converts retrospective enforcement information and public audit knowledge into prospective, evidence-admissible supervision.

\subsection{Stage 2: Evidence-Closed Audit Alignment}
\label{sec:stage2}
\begin{figure}[!t]
    \centering
    \includegraphics[width=\linewidth]
    {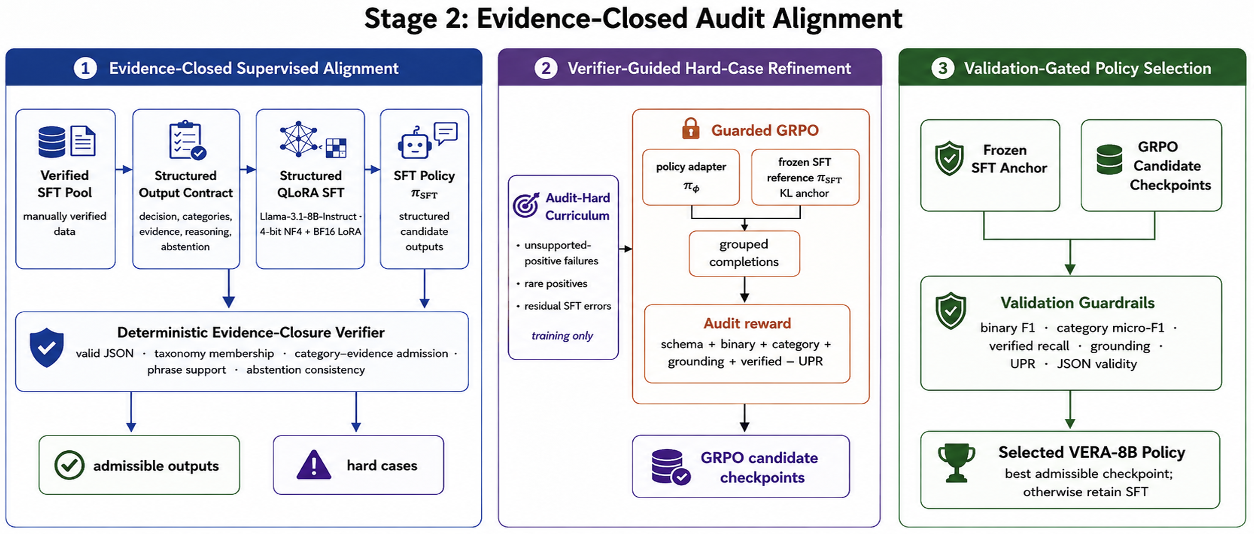}
    \caption{Stage 2: Evidence-admitted targets teach the contract, a verifier governs generation and refinement, and calibrated deferral routes uncertain judgments.}
    \Description{Stage 2 pipeline connecting evidence-closed supervised learning, a shared executable verifier, adaptive residual GRPO, and evidence-qualified selective routing.}
    \label{fig:stage2}
\end{figure}

Building on Stage-1 evidence standards, Stage 2 translates them into learnable, verifiable behavior. We apply three coupled mechanisms: evidence-closed contract learning, audit-constrained adaptive residual refinement, and evidence-qualified selective routing. A shared verifier enforces the same output contract across all three. Together, these mechanisms enforce one audit-evidence standard across learning, generation, verification, refinement, and deferral.
\paragraph{Mechanism 1: Evidence-closed contract learning.}
Mechanism~1 teaches the model the same evidence contract constructed in Stage~1. Let \(\mathcal{U}=\mathcal{P}\cup\mathcal{C}\) denote the strict positive observations and contamination screened controls constructed in Stage 1. A positive category enters supervision only when at least one contemporaneous filing span passes its category-specific evidence gate:
\begin{equation}
\forall k\in\mathcal{Y}_i^\star,\qquad \exists e\in E_i^\star \ \text{such that}\ G_k(e)=1.
\label{eq:supervision-admission}
\end{equation}
If a filing lacks admissible contemporaneous evidence, we do not label it as positive. This applies equally to clean observations, audit-hard negatives, and outcome-linked filings. The gate decouples enforcement outcomes from audit evidence, since enforcement points to what deserves investigation but the gate regulates what the model may learn.

For model training, the Stage~1 target is serialized into the following gold contract:
\begin{equation}
o_i^\star
=
\left(\mathcal{F}_i^\star,c_i^\star,a_i^\star\right),
\quad
\mathcal{F}_i^\star
=
\left\{
\left(
k_{ij}^\star,
e_{ij}^\star,
\phi_{ij}^\star,
r_{ij}^\star
\right)
\right\}_{j=1}^{m_i}.
\label{eq:audit-contract}
\end{equation}
Each flag contains a risk category, an evidence identifier, an exact supporting phrase, and a mechanism-level rationale. The contract also carries a model-visible confidence field \(c_i^\star\) and an abstention indicator \(a_i^\star\). Multiple supported categories
produce multiple flags, while abstention produces an empty flag set.

After serialization \(u_i^\star=\operatorname{Serialize}(o_i^\star)\), structured QLoRA SFT minimizes the answer-only causal language-model loss
\begin{equation}
\mathcal{L}_{\mathrm{SFT}}(\theta)=-\frac{1}{N}\sum_{i=1}^{N}\frac{1}{|A_i|}\sum_{t\in A_i}\log p_\theta\left(u_{it}^\star\mid x_i,u_{i,<t}^\star\right),
\label{eq:sft}
\end{equation}
where \(A_i\) indexes the answer tokens; normalization by \(|A_i|\) prevents longer multi-flag outputs from dominating the loss. The model jointly learns category selection, evidence attribution, rationale generation, and abstention as one autoregressive task. This yields \textsc{VERA-SFT}.

\paragraph{Shared executable verifier.}
The generated sequence is parsed as \(\hat{o}_i=\operatorname{Parse}(\hat{u}_i)\) and is admissible only when
\begin{equation}
V_i=I_i^{\mathrm{parse}}I_i^{\mathrm{tax}}I_i^{\mathrm{id}}I_i^{\mathrm{quote}}I_i^{\mathrm{state}}=1.
\label{eq:verification}
\end{equation}
The validator enforces structural compliance: valid JSON, taxonomy-consistent categories, and evidence identifiers present in the input. It further requires literal phrase containment after normalization and consistency between abstention and flag set ($\hat{a}_i=1$ only when $\hat{\mathcal{F}}_i=\varnothing$). This verifier governs the entire pipeline, keeping the same evidence standard from SFT through refinement and final routing.

\paragraph{Mechanism 2: Audit-Constrained Adaptive GRPO Refinement.}
Mechanism~2 uses GRPO to correct the residual errors left by \textsc{VERA-SFT}. The supervised model already satisfies the evidence contract but still misses some secondary risks and produces occasional unsupported positives. Adaptive refinement targets these hard cases while limiting changes to well-learned behaviors. For prompt \(i\), candidate completion \(j\) receives
\begin{equation}
\begin{aligned}
R_{ij}={}&
\lambda_s R_{ij}^{\mathrm{schema}}
+\lambda_b R_{ij}^{\mathrm{binary}}
+\lambda_c R_{ij}^{\mathrm{category}} \\
&+\lambda_e R_{ij}^{\mathrm{evidence}}
+\lambda_v R_{ij}^{\mathrm{verified}}
+\lambda_a R_{ij}^{\mathrm{asymmetry}},
\end{aligned}
\label{eq:reward}
\end{equation}
The reward combines six audit-specific objectives. \(R^{\mathrm{schema}}\) checks validity, \(R^{\mathrm{binary}}\) scores the positive/abstain decision, and \(R^{\mathrm{category}}\) scores multi-label recovery. \(R^{\mathrm{evidence}}\) rewards exact grounding, \(R^{\mathrm{verified}}\) requires the category--evidence pair to pass the verifier, and \(R^{\mathrm{asymmetry}}\) assigns audit-specific costs to unsupported positives and missed supported risks.

GRPO learns from reward differences within each candidate group. Once an objective is well learned, it provides little preference signal. Unresolved objectives still separate candidates. We use within-group reward variance to adapt each reward weight:
\begin{equation}
\begin{aligned}
\widetilde{\lambda}_k(t)
&=\lambda_k^{(0)}
\frac{\operatorname{Var}_k(t)+\epsilon}
{\frac{1}{K}\sum_{k'=1}^{K}
(\operatorname{Var}_{k'}(t)+\epsilon)},\\
\lambda_k(t)
&=\operatorname{clip}\!\left(
\gamma\lambda_k(t-1)
+(1-\gamma)\widetilde{\lambda}_k(t),
\lambda_{\min},\lambda_{\max}
\right).
\end{aligned}
\label{eq:variance-gate}
\end{equation}
Here \(\lambda_k^{(0)}\) is the base weight and \(\lambda_k(t)\) is the effective training weight. Exponential smoothing and clipping prevent short-run variance spikes from dominating optimization. This automatically shifts pressure from saturated objectives to those still capturing residual audit errors, without any manual reward schedule.

Policy updates remain anchored to the frozen \textsc{VERA-SFT}
policy:
\begin{equation}
\mathcal{L}_{\mathrm{GRPO}}
=
-\mathbb{E}\!\left[
\rho_{ijt}\widehat{A}_{ij}
-\beta D_{ijt}^{\mathrm{KL}}
\right].
\label{eq:grpo}
\end{equation}
A KL penalty limits policy drift, and a validation gate blocks any checkpoint that trades unsupported claims for recall. Concretely, a checkpoint qualifies only if it improves \textsc{EV-Micro-F1} or verified-positive recall while holding binary F1, JSON validity, and evidence grounding, without increasing \textsc{UCR}. Therefore, SFT defines the audit contract and the adaptive GRPO only corrects residual errors within it.

\paragraph{Mechanism 3: Evidence-Qualified Uncertainty Routing.}
Mechanism~3 controls the transition from model prediction to audit action. It converts model uncertainty into three actions: automatic positive, automatic negative, or review. Evidence remains mandatory for every automatic positive. We measure the model's preference for a positive finding over abstention from the likelihood of the two structured decision states:
\begin{equation}
m_i
=
\log
\frac{p_{\theta}(a_i=0\mid x_i)}
{p_{\theta}(a_i=1\mid x_i)},
\qquad
p_i=\operatorname{Cal}(m_i).
\label{eq:decision-margin}
\end{equation}
The calibration map \(\operatorname{Cal}(\cdot)\) is fit only on the disjoint calibration split. The resulting score measures uncertainty between a positive finding and abstention.

We then construct a split-conformal binary prediction set \(\mathcal{C}_i\subseteq\{0,1\}\) and combine it with the same evidence verifier used during training.
Let
\(A_i \equiv
[\mathcal{C}_i=\{1\}]
\land[V_i=1]
\land[\forall k\in\widehat{\mathcal F}_i,\exists e:G_k(e)=1]\)
denote an evidence-admissible positive, and
\(N_i \equiv
[\mathcal{C}_i=\{0\}]
\land[V_i=1]
\land[\widehat{\mathcal F}_i=\varnothing]\)
a valid negative. We then route
\begin{equation}
\mathrm{route}(i)=
\begin{cases}
\mathrm{Auto{+}}, & A_i,\\
\mathrm{Auto{-}}, & N_i,\\
\mathrm{Review}, & \text{otherwise}.
\end{cases}
\label{eq:selective-route}
\end{equation}
Only evidence-admissible predictions are automated. Ambiguous, malformed, or evidence-incomplete outputs are sent to review. 

Routing and certification are separate. The routing rule is fixed on a disjoint calibration set before testing, while certification additionally requires a finite-sample bound on false admissions. If the bound fails, we report routing diagnostics without a formal certificate. Together, these controls extend the audit-evidence standard from prediction to deployment.

\section{Experiment Results}
\label{sec:experiment-result}


\subsection{Experimental Setup and Evaluation Protocol}
\label{subsec:protocol}

We evaluate the full audit contract under one locked, issuer-disjoint protocol. We use 1,960 verified filings, divided by issuer into training (1,352), validation (298), and a frozen test set (310), with no issuer shared across splits. Validation is further separated by issuer into 199 observations for model selection and 99 for uncertainty calibration. All inputs are restricted to pre-enforcement filing text. We exclude cross-split exact and near-duplicate passages, event-specific AAER narratives, remediation disclosures, and other post-enforcement information. The test set is opened only after the model and uncertainty procedure are locked.

Three groups of models are compared under this protocol: general-purpose instruction-tuned models, finance-specific reasoning models, and our own model. All models receive the same filing inputs and output schema and are evaluated by the same parser and deterministic evidence verifier. The frozen test set is evaluated only once. Comparisons between \textsc{VERA-SFT} and \textsc{VERA-8B} use 10,000 paired observation-level bootstrap resamples and an exact two-sided McNemar test.

We score every output as generated, with no post-hoc repair. Invalid and incomplete outputs count as failures. Then we report binary accuracy and F1, category micro-F1, supported-category macro-F1, exact-set match, Evidence-Verified Micro-F1 (\textsc{EV-Micro-F1}), verified-positive recall, JSON validity, and the Unverified Claim Rate (\textsc{UCR}). Grounding, exact-span agreement, and evidence-token F1 serve as diagnostics. Under this standard, a predicted category earns \textsc{EV-Micro-F1} credit only when its accompanying evidence passes the same admissibility gate used throughout the pipeline. \textsc{UCR} measures the fraction of substantive category claims that lack admissible evidence. Malformed predictions receive no credit and are not silently removed. These metrics evaluate the complete audit contract rather than binary prediction alone.


\begin{figure}[tbp]
    \centering
    \includegraphics[width=\linewidth]
    {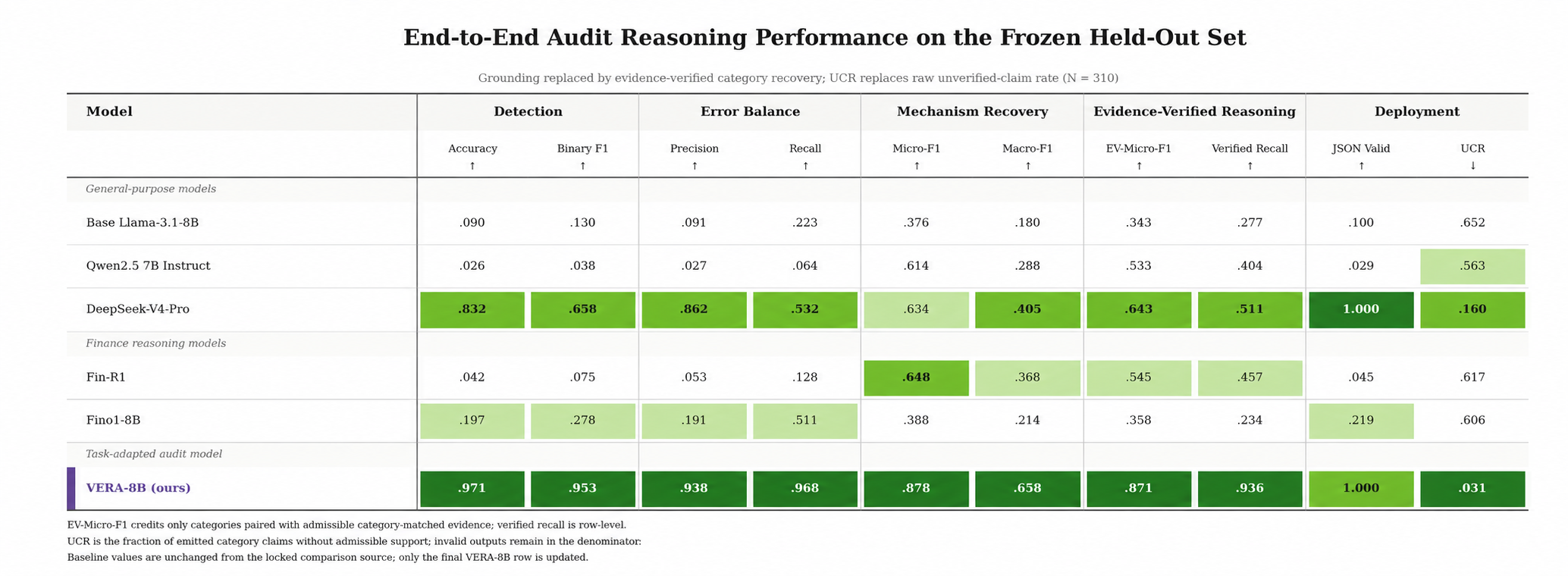}
    \caption{Frozen test-set comparison (\(N=310\)) of all models under
    strict unrepaired scoring. \textsc{EV-Micro-F1} requires admissible
    evidence, and \textsc{UCR} counts unsupported claims.}
    \Description{End-to-end comparison of detection, mechanism recovery,
    evidence-verified reasoning, and structural reliability.}
    \label{fig:model-comparison}
\end{figure}


\subsection{VERA-8B Leads the Full Audit Benchmark}
\label{subsec:end-to-end}

VERA-8B outperforms both finance-specific and general-purpose models by combining stronger audit-risk reasoning with verified evidence. Figure~\ref{fig:model-comparison} shows this advantage across both baseline families. Against the strongest finance-specific baseline, VERA-8B raises category micro-F1 from .648 to .878 and delivers substantially stronger evidence-verified reasoning. This finding indicates that financial knowledge alone is insufficient, and audit reasoning requires admissible evidence for each category. Compared with DeepSeek-V4-Pro, VERA-8B improves binary F1 by 45\% and verified-positive recall by 83\%, while reducing the unverified claim rate from 16.0\% to 3.1\%. This pattern shows that strong general reasoning does not by itself provide the evidence discipline required for prospective audit decisions. Our model succeeds by combining financial reasoning with the evidence discipline required for audit decisions.

This combined strength is reflected in VERA-8B's end-to-end performance. On the frozen test set, VERA-8B achieves 95.3\% binary F1 and 87.1\% \textsc{EV-Micro-F1}, with valid JSON for all 310 observations. These results show that its advantage extends beyond prediction accuracy to the full audit task, jointly supporting risk detection, multi-label reasoning, admissible evidence, and abstention.

\subsection{Audit-Specific Post-Training Drives the Main Capability Gains}
\label{subsec:post-training}

Our audit-specific post-training framework turns the base Llama-3.1-8B model into a new evidence-grounded audit reasoner. Structured SFT produces the main improvement from the base Llama-3.1-8B model, while adaptive GRPO further refines supported-risk recovery and verified category--evidence alignment. Starting from the same Llama-3.1-8B backbone, structured SFT raises binary F1 from 13.0\% to 94.7\% and \textsc{EV-Micro-F1} from 34.3\% to 86.4\%. It also produces valid JSON for all frozen-test observations and reduces the unverified claim rate from 65.2\% to 3.2\%. These gains, also summarized in Figure~\ref{fig:model-comparison}, show that structured supervision does more than improve prediction accuracy. It teaches the model to connect audit-risk decisions with admissible evidence and structured outputs under one audit-evidence contract.

Adaptive GRPO further improves the model on audit-specific errors. It focuses refinement on missed supported risks and verified category--evidence alignment without relaxing the evidence standard. Verified-positive recall rises from 91.5\% to 93.6\%, while false negatives fall from five to three. At the same time, JSON validity remains at 100\% and the unverified claim rate does not increase. This pattern indicates that adaptive refinement improves supported-risk
coverage without relaxing the evidence standard. The statistical tests also support this targeted interpretation. The paired 95\% bootstrap intervals include zero, and the exact McNemar test gives \(p=1.0\), consistent with GRPO refining specific audit errors rather than broadly changing the policy. These results support our unified post-training framework. More importantly, the results show that VERA-8B is an effective audit-specific reasoner and provide a practical foundation for evidence-grounded audit screening and review.

\subsection{Uncertainty Routing Directs Cases to Review}
\label{subsec:uncertainty}
\begin{table}[!ht]
\centering
\caption{Evidence-qualified routing of VERA-8B on the frozen test set
(\(N=310\)).}
\label{tab:routing}

\footnotesize
\renewcommand{\arraystretch}{0.94}
\setlength{\tabcolsep}{5pt}

\begin{tabular*}{0.92\linewidth}{
    @{\extracolsep{\fill}}
    lcc
}
\toprule
\textbf{Outcome / Diagnostic}
& \textbf{Cases}
& \textbf{Value} \\
\midrule

Automatic positive
& 76
& 24.5\% \\

Automatic negative
& 182
& 58.7\% \\

Manual review
& 52
& 16.8\% \\

\midrule

\textbf{Automatic coverage}
& 258
& \textbf{83.2\%} \\

\textbf{Automatic-negative NPV}
& --
& \textbf{100.0\%} \\

\textbf{Automatic-positive \textsc{UCR}}
& --
& \textbf{0.0\%} \\

Selective risk
& --
& 1.94\% \\

\bottomrule
\end{tabular*}
\end{table}
Uncertainty routing converts VERA-8B's predictions into a controlled audit-review policy. On the frozen test set, the system automatically handles 258 of 310 filings (83.2\%), leaving only 52 cases (16.8\%) for manual review. Table~\ref{tab:routing} shows that high automatic coverage is achieved without relaxing the evidence standard. VERA-8B automatically handles 83.2\% of the frozen test set, while all 182 automatic negatives achieve 100\% negative predictive value and automatically routed positive claims have zero \textsc{UCR}. The resulting selective risk is 1.94\%. In short, the routing layer separates confident, evidence-backed decisions from cases that require manual review.

Routing and certification remain separate by design. Certification is fail-closed, which means a deployment certificate is issued only when calibration evidence meets a high bar. The certification bound does not pass, so no certificate is issued. This ensures that weak evidence is never presented as a guarantee. Together, they turn model predictions into an evidence-qualified workflow that directs audit attention to the cases needing it most.

\begin{figure}[H]
    \centering
    \includegraphics[width=\linewidth]
    {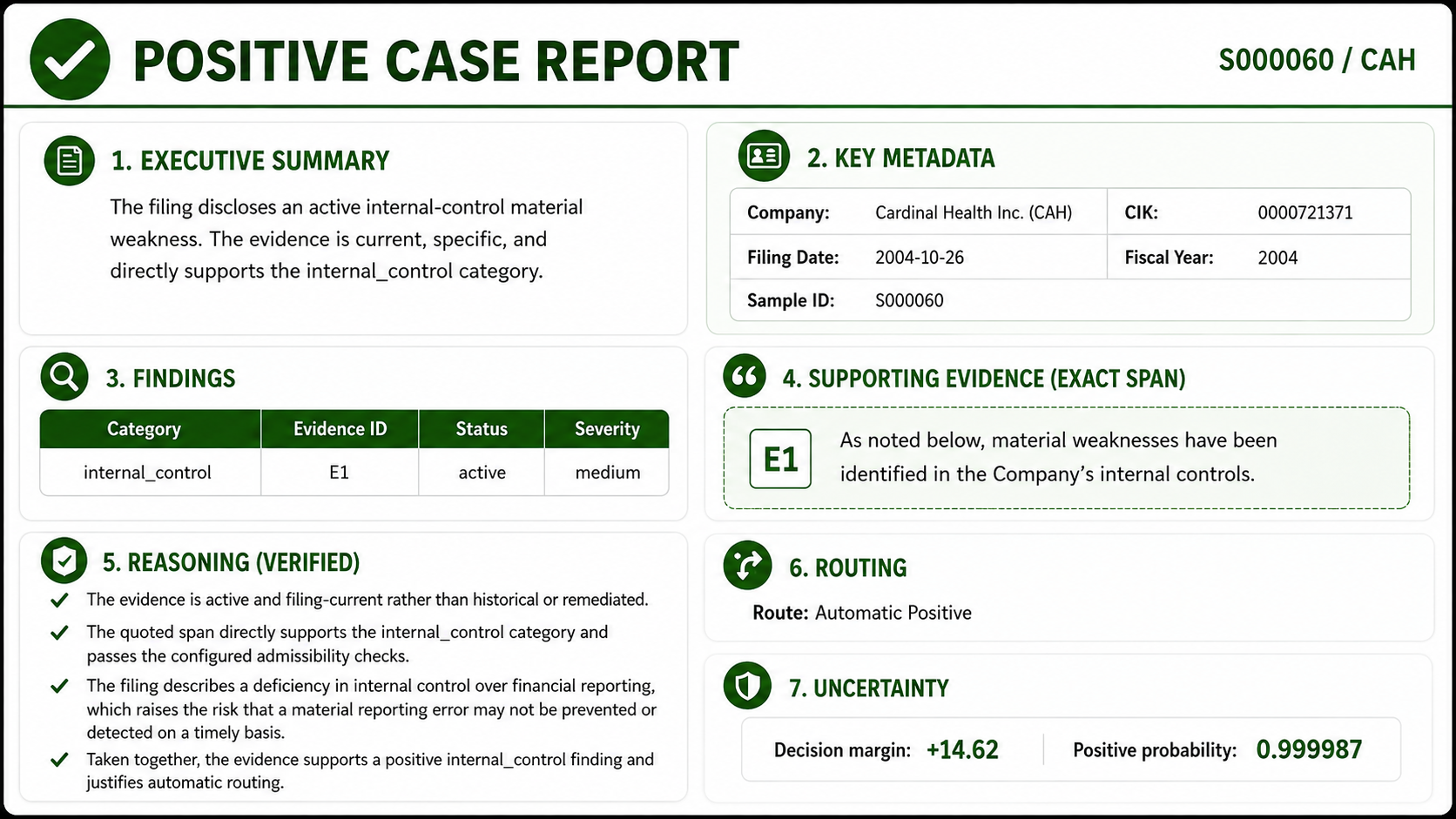}
    \caption{Reviewer-ready VERA-8B output for a frozen-test positive
    case.}
    \Description{A reviewer-ready positive case report presents the
    filing identifier, detected internal-control risk, exact supporting
    evidence, reasoning, routing status, and calibrated uncertainty.}
    \label{fig:practical-output}
\end{figure}
\subsection{VERA-8B Supports Practical Audit Review}
\label{subsec:practical-output}

We develop AuditBridge as a new output layer that connects VERA-8B's verified reasoning to both computational processing and audit review. Each decision is stored as a machine-readable JSON record and rendered as a reviewer-ready audit report. The JSON turns each decision into a structured record for downstream financial processing.

The reviewer-ready report turns the same verified record into a usable audit output. Figure~\ref{fig:practical-output} shows one frozen-test positive case, presenting the filing metadata, risk finding, exact supporting evidence, verified reasoning, routing decision, and uncertainty in one compact record. Together, these two output forms make VERA-8B practical for both machine processing and audit review, while keeping every decision traceable to verified filing evidence.

\section{Conclusion}

We introduce VERA-8B, an audit-specific reasoner that turns
pre-enforcement SEC filings into evidence-grounded risk findings.
A unified post-training framework teaches the model to identify risks,
cite admissible evidence, abstain when support is insufficient, and
refine audit-specific errors under one evidence standard. On the frozen
test set, VERA-8B leads all evaluated baselines while maintaining
structured outputs and low unsupported-claim rates. Uncertainty routing
directs ambiguous cases to review, and AuditBridge connects verified
outputs to downstream processing and practical audit review. VERA-8B
therefore provides a practical foundation for scalable,
evidence-grounded audit workflows.

\clearpage
\bibliographystyle{unsrtnat}
\bibliography{references}
\clearpage
\end{document}